\documentclass[11pt,a4paper]{article}
\usepackage[hyperref]{conll-2019}
\usepackage{times}
\usepackage{latexsym}
\usepackage{multirow}
\usepackage{graphicx}

\usepackage{url}

\aclfinalcopy 

\title{Interpretable Structure-aware Document Encoders \\ with Hierarchical Attention}

\author{Khalil Mrini \\
  Computer Science and Engineering \\
  University of California, San Diego \\
  La Jolla, CA 92093 \\
  {\tt \small Khalil@ucsd.edu} \\\And
  Claudiu Musat, Michael Baeriswyl \\
  AI Group \\
  Swisscom AG \\
  Lausanne, Switzerland \\
  {\tt \small firstname.lastname@swisscom.com}\And
  Martin Jaggi \\
  Machine Learning and \\
  Optimization Lab, EPFL \\
  Lausanne, Switzerland \\
  {\tt \small martin.jaggi@epfl.ch}}

\date{}

\begin{document}
\maketitle
\begin{abstract}
  We propose a method to create document representations that reflect their internal structure. We modify Tree-LSTMs to hierarchically merge basic elements such as words and sentences into blocks of increasing complexity. Our \textit{Structure Tree-LSTM} implements a hierarchical attention mechanism over individual components and combinations thereof. We thus emphasize the usefulness of Tree-LSTMs for texts larger than a sentence\footnote{GitHub Repository: \\ \url{https://github.com/swisscom/ai-research-document-classification}}.
  We show that structure-aware encoders can be used to improve the performance of document classification. We demonstrate that our method is resilient to changes to the basic building blocks, as it performs well with both sentence and word embeddings. The \textit{Structure Tree-LSTM} outperforms all the baselines on two datasets by leveraging structural clues\footnote{We release our Wikipedia dataset here: \\ \url{https://zenodo.org/record/2642190\#.XZgrPPlKhVo}}. We show our model's interpretability by visualizing how our model distributes attention inside a document.
  On a third dataset from the medical domain, our model achieves competitive performance with the state of the art. This result shows the \textit{Structure Tree-LSTM} can leverage dependency relations other than text structure, such as a set of reports on the same patient.
\end{abstract}

\section{Introduction}

Humans use structure to better represent information, and within that structure, elements vary in importance. For example, a table of contents helps in defining a document's global structure and focus the reader's attention to what matters.

Long, unstructured sequences are hard to process for humans and machines alike. Even though neural network techniques have recently shown significant improvement to text classification, Long Short-Term Memory (LSTM) networks \cite{hochreiter1997long} perform poorly for long sequences \cite{cheng2016long}. Proposed solutions to overcome the problems that stem from long, flattened sequences include LSTM variants such as the bidirectional variant \cite{graves2005framewise, huang2015bidirectional, chiu2015named} and the attention-based one \cite{wang2016attention}. The latter focuses on relevant sections independent of their location. 
Flat attention cannot, however, cope with long sequences. Splitting a long text into smaller sections has the advantage of being able to flush the attention when the local context ends. A first step in this direction has been taken by \citet{yang2016hierarchical}, who apply attention for words, and sentences as well, but documents are still viewed as a flat sequence of sentences.

Our first contribution is to adapt tools this far used only in the context of a single sentence, the tree-structured LSTM networks \cite{tai2015improved, le2015compositional, zhu2015long}, to fit the document structure instead.

Our second contribution is to extend previous attention mechanisms to all the structural levels of a document, and make them applicable in a hierarchical structure. To do so, we apply different initialization mechanisms to our Tree-LSTM model, leading to a transformation of the LSTM forget gates into a \textit{de facto} attention mechanism. We show our model is interpretable and makes semantically relevant choices using a visualisation of this hierarchical attention mechanism.

By including hierarchical structure in the document representation and leveraging several layers of visualizable attention, we create \textbf{interpretable structure-aware attention-based document encoders}. 





Our third contribution is to show that the structure-aware encoders are useful. We choose the task of supervised multi-class document classification as a first target, where each document has to be assigned to exactly one category. 
We first show that hierarchical documents can be better classified using tree-structured LSTMs. We obtain \textbf{improved document classification results} over two datasets with varying structure depths.

Then we show that our model can leverage structure beyond a single document, in settings where a training sample is a set of related documents. On a benchmark dataset from the medical domain, we model patients using their health reports to predict mortality. We obtain competitive results with the state of the art, emphasizing that our model can \textbf{efficiently model dependency relations other than simple textual structure.}

Finally, we also release our code along with our new dataset of hierarchical documents.

The rest of the paper is organised as follows: we summarize the literature on document embedding and classification in Section \ref{2}. Section \ref{3} describes the proposed method and Section \ref{4} details the experiments and results. We draw conclusions and outline suggestions for future work in Section \ref{5}.

\section{Related Work}
\label{2}

\noindent \textbf{Text Embeddings.} Among the most popular context-based word embeddings models are \textit{word2vec} \cite{mikolov2013efficient, mikolov2013distributed}, \textit{FastText} \cite{bojanowski2016enriching} and \textit{GloVe} \cite{pennington2014glove}. 

More recently, embedding methods for sentences have emerged. \citet{zhao2015self} introduce \textit{AdaSent}, a self-adaptive hierarchical sentence model. \citet{kiros2015skip} propose \textit{Skip-Thought}: an extension of the skip-gram model to sentences. \citet{hill2016learning} try to address computationally expensive training with their \textit{FastSent} model. \citet{arora2016simple} compute a sentence's embedding as the average of the embeddings of its words, minus the first principal component. \citet{conneau2017supervised} propose a supervised sentence embedding model trained on the SNLI dataset \cite{bowman2015large}. The \textit{sent2vec} embeddings \cite{pagliardini2017unsupervised} are trained unsupervised, and represent sentences by looking at unigrams, as well as n-grams that compose them, in a similar fashion to FastText. \citet{devlin2018bert} introduce the Bidirectional Encoder Representations from Transformers (\textit{BERT}).

\noindent \textbf{Document Encoders.} \citet{yang2016hierarchical} introduce a hierarchical document encoder, used as a document classification model. It decomposes a document into a sentence and a word level. At each level, it applies an encoder with a bi-GRU and an attention mechanism. As a higher level structure is not used, they tested the method on shorter documents including reviews and answers.

\noindent \textbf{Tree-structured LSTM networks.} \citet{tai2015improved} adapt the standard LSTM to tree-like structures with two kinds of \textit{Tree-LSTM} architectures. The first one is the Child-Sum variant: for a unit $j$, the hidden variable $h_{j-1}$, that would be carried on from the previous LSTM unit $j-1$ in a standard architecture, is replaced by the sum of the hidden variables of its children units $\tilde{h}_j = \sum_{k \in C(j)}h_k$, with $C(j)$ being the children units of unit $j$. In addition, there is one forget gate $f_{jk}$ per child $k$ of unit $j$. The parameter matrices enable the unit to determine the contributions of its children units in each gate. The second variant is the $N$-ary Tree-LSTM, where each non-leaf unit should have a branching factor of at most $N$ and have ordered children. This variant allocates one parameter matrix per child, enabling it to learn conditioning based on the child's position from $1$ to $N$. However, it is not as modular as the Child-Sum Tree-LSTM as there is a constraint on the branching factor. Any Tree-LSTM unit still has to get an input $x$, whether it is a leaf unit or not.

\citet{le2015compositional} develop the \textit{LSTM-RNN}, a binary tree-structured LSTM architecture, such that each non-leaf unit has exactly two children, with the corresponding pairs of input and forget gates. \citet{zhu2015long} introduce a similar binary tree-structured LSTM architecture, the \textit{S-LSTM}, in which there is one input gate per non-leaf unit, but still two forget gates. A non-leaf unit in these two architectures does not have an input of its own, but it takes the \textit{inputs} of its children units for the \textit{LSTM-RNN} model, and their \textit{outputs} for the \textit{S-LSTM} model.


In addition to sentence-level sentiment classification, \citet{tai2015improved} test their model on semantic relatedness between sentence pairs. \citet{eriguchi2016tree} extend the Tree-LSTM model to introduce a tree-to-sequence attentional neural machine translation model. \citet{chen2017enhanced} use it in conjunction with a bi-LSTM to form a hybrid model for natural language inference.

\section{Structure-aware Attention-based Document Encoders}
\label{3}

\subsection{Structure Awareness}

\begin{figure}
\centering
\includegraphics[width=\columnwidth]{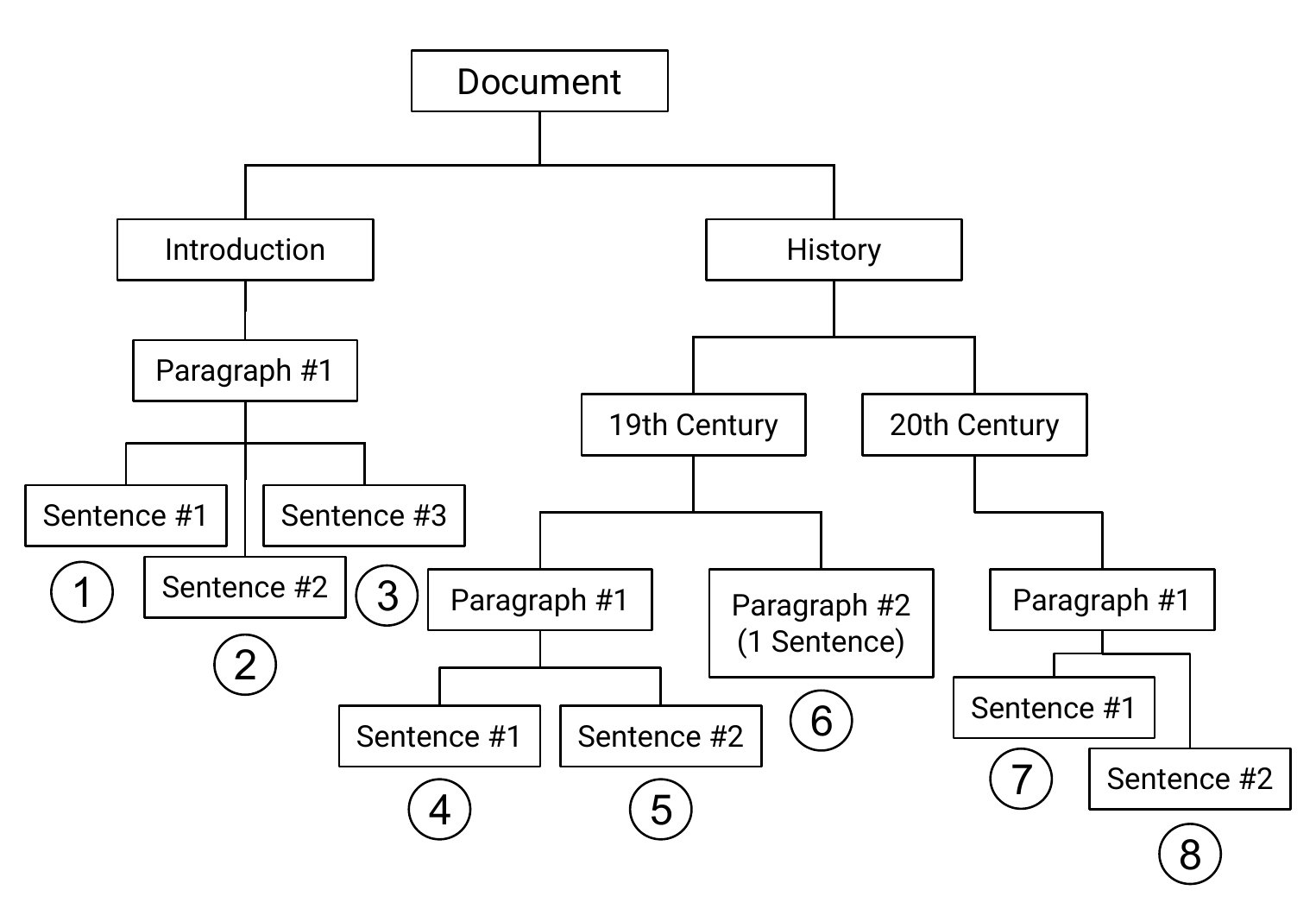}
\caption{The \textit{Structure Tree} corresponding to a document: the basis to form the document's Tree-LSTM.}
\label{fig1}
\end{figure}

\begin{figure}
\centering
\includegraphics[width=\columnwidth]{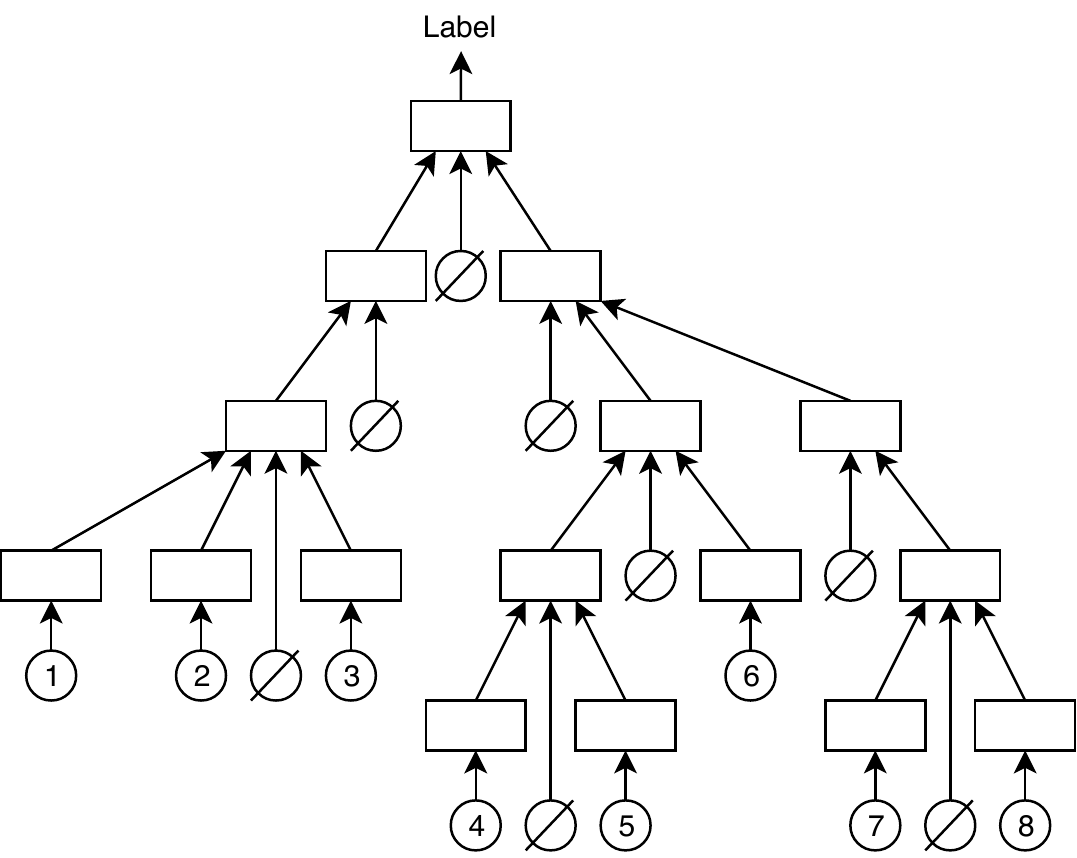}
\caption{The \textit{Structure Tree-LSTM with Zero Vectors} applied to the document in Figure \ref{fig1}, with the numbers in circles referring to the same input. The empty set symbols indicate a vector of zeros.}
\label{fig2}
\end{figure}

\begin{figure}
\centering
\includegraphics[width=\columnwidth]{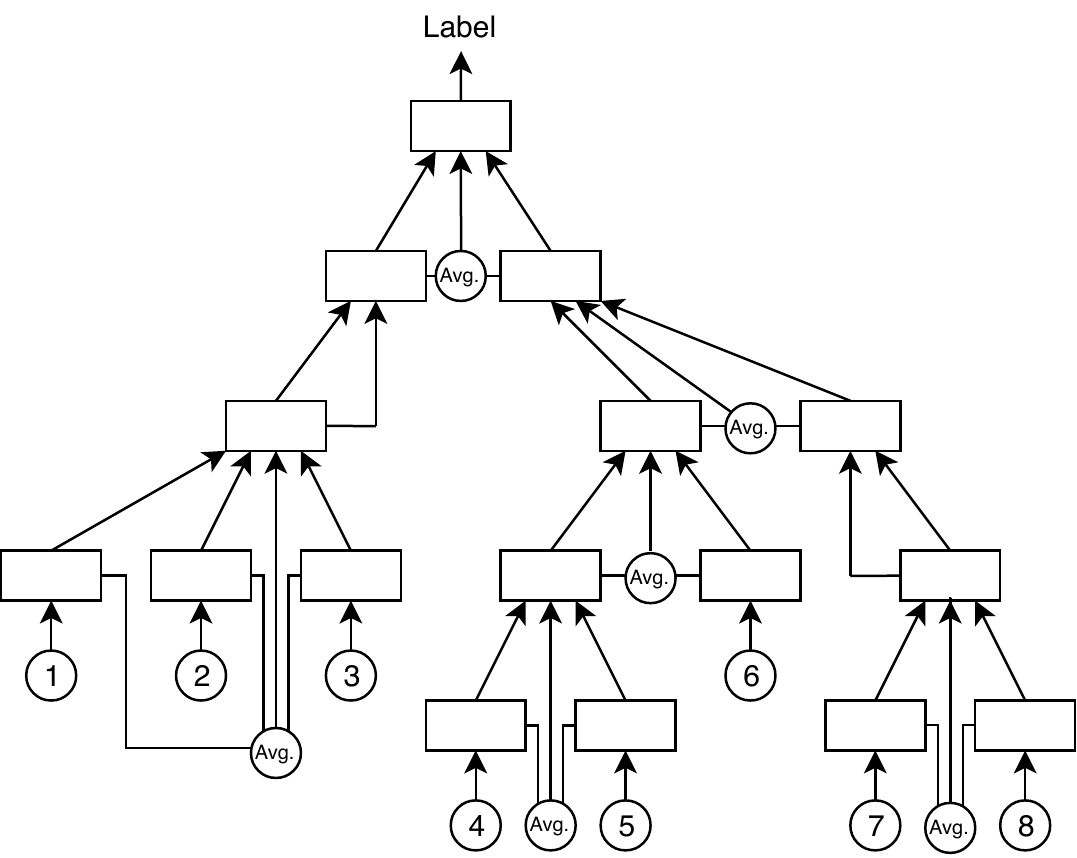}
\caption{The \textit{Structure Tree-LSTM with Hierarchical Average} applied to the document in Figure \ref{fig1}, with the numbers in circles referring to the same input. A given unit's input is the average of the inputs of its children.}
\label{fig3}
\end{figure}

The starting assumption is that common documents have a hierarchical structure. Words are grouped in sentences, sentences in paragraphs, which in turn form subsections, sections and so on. 
From this observation, we derive the hypothesis that hierarchical attention over a document's structure allows the resulting representation to highlight the document's important aspects. 

Our \textit{Structure Tree-LSTM} captures a document's hierarchical structure by mirroring the corresponding \textit{document tree}. For example, the document tree in Figure \ref{fig1} corresponds to a document with the following outline: (1) \textit{Introduction}: 1 paragraph with 3 sentences; (2) \textit{History}: 2 subsections; (2.1) \textit{19th Century}: 2 paragraphs, with 2 and 1 sentences respectively; (2.2) \textit{20th Century}: 1 paragraph with 2 sentences.
The structure granularity can be adjusted according to the downstream task and size of the dataset. Large datasets can have coarse-grained structure for the model to be less computationally expensive, whereas smaller datasets can have Tree-LSTMs include components all the way down to words.


The first major difference with respect to the existing Tree-LSTM, is that, as seen in the example, there is no imposed order on the semantic components. The \textit{Dependency Tree-LSTM} of \citet{tai2015improved} uses sentence-level word dependencies, as in the example in Figure \ref{fig4}. This is generally not extensible at the document level. Our \textit{Structure Tree-LSTM} relaxes this assumption, making it more generally applicable. 

The second major difference is the distinction between leaf and non-leaf units in a \textit{Structure Tree-LSTM}.
A leaf unit is the smallest component of the document (i.e., a word or a sentence), and has as input the component's embedding. A non-leaf (parent) unit represents a larger component of the document: a sentence (group of words), a paragraph (group of sentences), or a section (group of sections and/or paragraphs). This generalization of the node contents allows for the extension of the method to more general contexts. In the original models of \citet{tai2015improved}, all units of a Tree-LSTM represent an original input (i.e., a word).

This distinction between unit types leads to the creation of two variants of \textit{Structure Tree-LSTMs}, that differ in the strategy for filling the non-leaf units.
We investigate two main strategies:

\begin{figure}
\centering
\includegraphics[width=3cm]{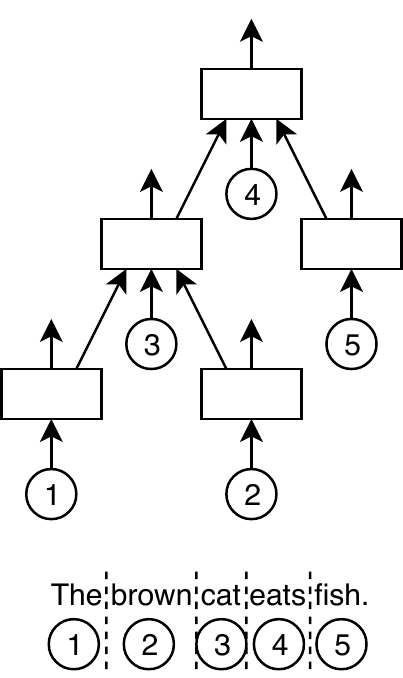}
\caption{Example of a sentence encoding using the \textit{Dependency Tree-LSTM} of \citet{tai2015improved}. Each LSTM unit receives a word embedding as input. This sentence's reordering is based on its dependency tree.}
\label{fig4}
\end{figure}

\textbf{(1)} \textbf{Structure Tree-LSTM with Zero Vectors}: non-leaf units get zero vectors as input (Figure \ref{fig2}).

\textbf{(2)} \textbf{Structure Tree-LSTM with Hierarchical Average}: non-leaf units have as input the average of the input vectors of its children (Figure \ref{fig3}).

The \textit{Structure Tree-LSTM} is easily extensible with additional initialization methods. One such method could be replacing the hierarchical averaging by a sum of the children's inputs. Another one could be using section titles as input for the non-leaf units representing sections. This underlines the power of the \textit{Structure Tree-LSTM} to incorporate all the information available. However, for the sake of fairness in comparison with the baselines, we did not use section titles in any of our models.

\subsection{Hierarchical Attention}

We use the same transition equations as in the Child-Sum Tree-LSTM described in \citet{tai2015improved}. We analyse them to explain the attention mechanisms of our proposed models.

For a unit $j$ of a Child-Sum Tree-LSTM, the hidden state $h_{j-1}$, that is carried on from the previous LSTM unit $j-1$ in a standard architecture, is replaced by the sum of the hidden states of its children units $\tilde{h}_j = \sum_{k \in C(j)}h_k$, with $C(j)$ being the children units of unit $j$. In addition, there is one forget gate $f_{jk}$ per child $k$ of unit $j$. 

However, as the leaf units have no children units, we have that $\left|C(j)\right| = 0$, and as such $\tilde{h}_j = 0$. Therefore, the only contribution comes from the input (the word or sentence embeddings), without influence from other inputs. This changes the equations in practice, as for example the formula for the \textit{input} gate:

\begin{equation}
i_j = \sigma \left(W^{(i)}x_j + U^{(i)}\tilde{h}_j + b^{(i)} \right)
\label{eq2}
\end{equation}
becomes for leaf units:

\begin{equation}
i_j = \sigma \left(W^{(i)}x_j + b^{(i)} \right)
\label{eq21}
\end{equation}

The model with \textbf{Zero Vectors} \textit{de facto} changes 
the formulas for the non-leaf units as well. Because a non-leaf unit's input is $x_j = 0$, the only contribution comes from the children units. This makes the \textit{Structure Tree-LSTM} with zero vectors similar to a joint hierarchical attention mechanism. The formula for the \textit{forget} gate for child unit $k$:

\begin{equation}
f_{jk} = \sigma \left(W^{(f)}x_j + U^{(f)}h_k + b^{(f)} \right)
\label{eq3}
\end{equation}
becomes for non-leaf units:

\begin{equation}
f_{jk} = \sigma \left(U^{(f)}h_k + b^{(f)} \right)
\label{eq31}
\end{equation}
making the \textit{forget} gate an attention mechanism over individual child units. Likewise, the formulas of the \textit{memory cell} $c_j$, the \textit{input} gate $i_j$ and \textit{output} gate $o_j$ change in practice. For example, the formula for the \textit{output} gate:
\begin{equation}
o_{j} = \sigma \left(W^{(o)}x_j + U^{(o)}\tilde{h}_j + b^{(o)} \right)
\label{eq4}
\end{equation}
becomes for non-leaf units:
\begin{equation}
o_{j} = \sigma \left(U^{(o)}\tilde{h}_j + b^{(o)} \right)
\label{eq41}
\end{equation}
such that the \textit{output} gate can now be assimilated to an attention over linear combinations of individual children. 

We can thus view the model with \textbf{Zero Vectors} as a generalization of hierarchical attention mechanisms.



\section{Experiments}
\label{4}

\subsection{Experimental Setup} 

We design two document classification experiments and one mortality prediction experiment for the \textit{Structure Tree-LSTM}.  
To evaluate our model, we focus the analysis on datasets having a hierarchical structure. Therefore, we could not use the datasets of reviews on which the Hierarchical Attention Networks \cite{yang2016hierarchical} are evaluated. 


In the datasets of reviews in \citet{yang2016hierarchical}, a training sample is only one paragraph with about 5 to 14 sentences on average. In our three selected datasets, we keep all information on internal structure: where each document part (paragraphs, sections, subsections...) starts and ends. We therefore evaluate our models on datasets with at least three levels of hierarchy, with the highest hierarchy level being larger than a paragraph. More concretely, the highest hierarchy level in our datasets are an article, an email or a patient's medical record.

\subsection{Text Structure in Document Classification}

\subsubsection{Document Classification Datasets}

\noindent \textbf{The Enron Email Dataset.} This UC Berkeley-labelled dataset\footnote{Collected from: \url{http://bailando.sims.berkeley.edu/enron_email.html}} contains 1,700 tagged emails with many overlapping categories. 
We select the two most common categories, and assign a third label for their intersection, and a fourth label for the emails that do not belong to any category.
These real-world documents present meaningful, yet minimalist structure expressed as paragraphs.

\noindent \textbf{The Wikipedia Dataset.} We collect an English Wikipedia dataset of 494,657 articles. These are relatively long articles, split into 24 disjoint categories and released as an open resource. More information is available in Appendix A. Given that this is a large dataset, we set leaf nodes to represent sentences to mitigate computational complexity: leaf nodes take \textit{sent2vec} embeddings as input.

\subsubsection{Baselines} 
For the Enron Email and Wikipedia datasets, we compare the \textit{Structure Tree-LSTM} with the following baselines:

\textbf{(3)} \textbf{MLP with Unweighted Average}: a Multi-Layer Perceptron (MLP) with one hidden layer having a rectifier activation function. In this model, a document is represented by an unweighted average of all input embeddings.

\textbf{(4)} \textbf{MLP with Hierarchical Average}: same architecture as model \textbf{(3)}, but the document representation is a hierarchically-weighted average of input embeddings.

\textbf{(5)} \textbf{Sequential LSTM}: an LSTM layer that takes input embeddings in sequential order. The sets of input embeddings are not padded nor truncated. This model shares the same input as the Tree-LSTM, but processes them sequentially rather than hierarchically.

\textbf{(6)} \textbf{Hierarchical Attention Networks}: the model\footnote{Code: \url{https://github.com/EdGENetworks/attention-networks-for-classification}} designed by \citet{yang2016hierarchical}. To remain faithful to the original model, this model is only tested on experiments with \textbf{word embeddings} as inputs. This model uses bidirectional GRUs, making it a hierarchical bidirectional RNN model. The model's hierarchy has two levels: one for words and one for sentences. Each level has separate encoders and attention weights, as well as a fixed number of elements. This means there is a fixed number of words (resp. sentences) per sentence (resp. document), and as such padding or truncating are applied where necessary.

We compare the number of parameters for each model in Table \ref{compl}. We take into account the hidden layer, as well as the \textit{softmax} output layer. Our \textit{Structure Tree-LSTM} models have as many parameters to learn as a sequential LSTM. Ignoring the output layer and the HAN attention weights, an MLP model has about 4 times less parameters to learn than an LSTM model, and the HAN model has about 3 times more parameters.

\subsubsection{Document Classification Results}

We evaluate the Enron Email and Wikipedia datasets using the Macro-F1 score, computed by averaging the individual F1 scores of each class. These two datasets are about multi-class document classification, and the Macro-F1 score takes into account class imbalance. We show our results in Table \ref{exp2}.

For the Enron Emails, the \textit{Structure Tree-LSTM} model with zero vectors obtains the highest F1 scores, outperforming all other models. The model's performance is resilient regardless of the kind of building block the leaves represent -- words or sentences. The extreme difference in macro-F1 scores with respect to the base LSTM underlines the importance of structure when fewer data points are available. We note the lacklustre macro-F1 of the Hierarchical Attention Networks, that suggests our structure-oriented attention model requires less data to train.

Likewise, for the Wikipedia dataset, the two \textit{Structure Tree-LSTM} models score visibly higher in both macro-F1 and accuracy than the baselines, confirming the efficiency of a document embedding inclusive of structure. The absolute gain is higher than for the previous dataset, suggesting that our models successfully leverage the additional structure in Wikipedia articles. In all cases, the best \textit{Structure Tree-LSTM} variant is the one with zero vectors, showing that attention over children units 
is sufficient.


We analyse the classification errors of the \textit{Structure Tree-LSTM} with Zero Vectors. The predictions of the \textit{actors} category are correct 83.12\% of the time. The two most common incorrect predictions for \textit{actors} are \textit{actresses} (3.89\%) and \textit{directors} (2.82\%). Another example is the \textit{airlines} category, with 85.54\% accuracy and most commonly confused with \textit{aircraft} (12.90\%). Although our model has high accuracy, its errors seem to stem from confusing semantically related categories. 

\begin{table*}[t!]
\begin{center}
\begin{tabular}{|l|l|}
\hline \bf Model & \bf Number of Parameters to Learn \\
\hline Structure Tree-LSTM & $4 ( e h + h^2 + 2h) + h l + l $ \\
\hline Sequential LSTM & $4 ( e h + h^2 + 2h) + h l + l $ \\
\hline MLP & $eh + h^2 + 2h + hl +l$ \\
\hline Hierarchical Attention Networks (HAN) & $12(eh +h^2 +2h)+w+s+sl+l$ \\
\hline 
\end{tabular}
\end{center}
\caption{\label{compl} Number of parameters to learn for each model in the document classification datasets. $e$ is the embedding dimension, $h$ is the hidden layer dimension, $l$ is the number of labels. For the HAN model, $w$ is the number of words per sentence and $s$ is the number of sentences per document.}
\end{table*}

\begin{figure*}
    \centering
    \includegraphics[width=11.8cm]{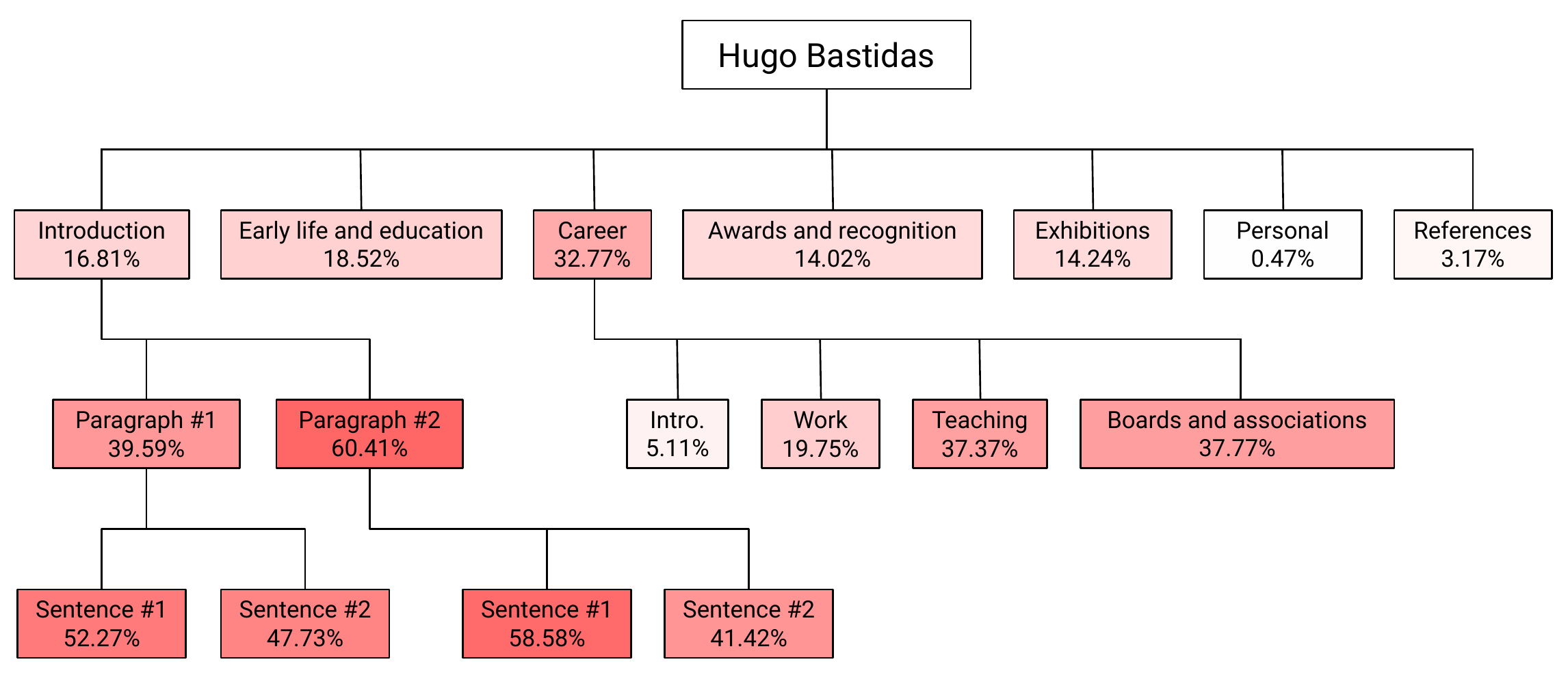}
    \caption{Visualisation of the hierarchical attention mechanism of our \textit{Structure Tree-LSTM} with Zero vectors. The example is the Wikipedia article of \textit{Hugo Bastidas}, correctly predicted as an artist. All cells, except the top one, are colored in the same red, with the opacity toned to the corresponding attention weights.}
    \label{ha}
\end{figure*}

\begin{table*}[t!]
\begin{center}\vspace{-1mm}
\begin{tabular}{|p{1.5cm}|p{1.9cm}|l|r|r|}
\hline \bf Dataset & \bf Leaves & \bf Model & \bf Macro-F1 & \bf Accuracy \\ \hline
\multirow{ 11}{1.6cm}{Enron Emails} & \multirow{ 6}{2cm}{Word Embeddings \textit{(word2vec)}} & Structure Tree-LSTM with Zero Vectors & \bf 0.4455 & \bf 0.5235 \\
& & Structure Tree-LSTM with Hierarchical Average & 0.4099 & 0.4824 \\ \cline{3-5}
& & MLP with Unweighted Average & 0.4063 & 0.4529 \\
& & MLP with Hierarchical Average & 0.3934 & 0.4941 \\
& & Sequential LSTM & 0.3429 & 0.4176 \\
& & Hierarchical Attention Networks & 0.3632 & 0.5078 \\ \cline{2-5}
& \multirow{ 5}{2cm}{Sentence Embeddings \textit{(sent2vec)}} & Structure Tree-LSTM with Zero Vectors & \bf 0.4533 & \bf 0.5118 \\
& & Structure Tree-LSTM with Hierarchical Average & 0.4278 & 0.4941 \\ \cline{3-5}
& & MLP with Unweighted Average & 0.4164 & 0.4588 \\
& & MLP with Hierarchical Average & 0.3822 & 0.4706 \\
& & Sequential LSTM & 0.3002 & 0.4059 \\ \hline
\multirow{ 5}{1.6cm}{Wikipedia} & \multirow{ 5}{2cm}{Sentence Embeddings \textit{(sent2vec)}} & Structure Tree-LSTM with Zero Vectors & \bf 0.8538 & \bf 0.8877 \\
& & Structure Tree-LSTM with Hierarchical Average & 0.8430 & 0.8814 \\ \cline{3-5}
& & MLP with Unweighted Average & 0.7870 & 0.8534 \\
& & MLP with Hierarchical Average & 0.7790 & 0.8476 \\
& & Sequential LSTM & 0.6405 & 0.7802 \\
\hline
\end{tabular}\vspace{-1mm}
\end{center}
\caption{\label{exp2} Results of the multi-class classification experiments. The best scores are in bold.}
\end{table*}

\begin{table*}[t!]
\begin{center}\vspace{-1mm}
\begin{tabular}{|p{1.7cm}|p{1.9cm}|p{6.4cm}|r|}
\hline \bf Dataset & \bf Leaves & \bf Model & \bf AUC score \\ \hline
\multirow{ 4}{2cm}{MIMIC-III} & \multirow{ 4}{2cm}{Sentence Embeddings \textit{(sent2vec)}} & Structure Tree-LSTM with Zero Vectors & 0.958 \\ \cline{3-4}
& & LDA & 0.930 \\
& & doc2vec & 0.930 \\
& & CNN & \bf 0.963 \\
\hline
\end{tabular}\vspace{-1mm}
\end{center}
\caption{\label{exp45} Results of the binary classification experiment in comparison with the baselines of \citet{grnarova2016neural}.}
\end{table*}

\subsubsection{Analysis of the Hierarchical Attention Mechanism}

To visualize the attention weights given by a node to its children nodes, we compute the product of each child's candidate vector with the forget gate. Then, we divide the value for each dimension of the resulting vector by the sum of the values for all children for the same dimension. We therefore obtain, for each child, a vector of contribution percentages for each dimension. We average this vector, and get the individual attention weights of the children nodes. More formally, for the $k$-th child of node $j$, with the child's candidate vector $c_k$ of dimension $D$ and corresponding forget gate $f_{jk}$, we compute the corresponding attention weight $a_{jk}$ as follows:

\begin{equation}
    a_{jk} = \frac{1}{D}\sum_{d=1}^{D}\frac{f_{jk}[d]*c_k[d]}{\sum_{k'}f_{jk'}[d]*c_{k'}[d]}
\end{equation}

We analyse how our \textit{Structure Tree-LSTM} with Zero vectors applies its hierarchical attention mechanism to documents through an example: the Wikipedia article of \textit{Hugo Bastidas}\footnote{Available here: \url{https://en.wikipedia.org/wiki/Hugo_Bastidas}}, correctly predicted by our model as an article about an artist. The attention weights are visualized in Figure \ref{ha}.

Our \textit{Structure Tree-LSTM} displays semantically relevant choices in its distribution of attention: to predict that the article is about an artist, it pays more attention to the \textit{Career} section. The figure shows that a model needs more than just the introduction of a document for efficient classification. We illustrate how our model attributes attention weights to all levels of the hierarchy, including subsections (under \textit{Career}), and all the way down to paragraphs and sentences (under \textit{Introduction}). We did not display all levels of the hierarchy for lack of space. Using this attention visualisation, we can interpret our model, and see which parts of the document influences the prediction.

\subsection{Extended Structure: Modelling Sets of Documents}

We use a dataset from the medical domain, where the documents are medical reports of different categories. The reports can be grouped by patient to form sets of documents modelled as one \textit{Structure Tree-LSTM}. By comparing to baseline models, we inquire whether our model can leverage additional structural knowledge, such as the types of reports and links between them.

\subsubsection{The MIMIC-III Dataset}

To compare our model to existing benchmark datasets, we use the MIMIC-III dataset \cite{johnson2016mimic}. It is a freely accessible database on critical medical care.

It contains over 2 million unstructured textual medical reports corresponding to 46,520 hospitalised patients, and information about whether a patient has eventually recovered. In case the patient died, it is indicated whether the death occurred at the hospital, within one month after leaving hospital care, or within a year afterwards.

\citet{grnarova2016neural} use this dataset to predict patient mortality. 
It contains 31,244 patients with 812,158 notes. \citet{grnarova2016neural} approach this task as multiple binary classification problems: to predict mortality during the hospital stay, within one month later, or within a year later. We obtain the filtered dataset from \citet{grnarova2016neural}, and focus on in-hospital mortality prediction.

In this experiment, the root node of the Structure Tree represents a patient, and its children units are the patient's reports.  Each of the reports are divided into paragraphs and sentences. Similarly to the Wikipedia dataset, we use sentence embeddings for this experiment. Like \citet{grnarova2016neural}, we implement our \textit{Structure Tree-LSTM} with categories encoded as one integer appended to sentence embeddings.

The training time for one epoch is over 24 hours. We choose to focus on the Zero-vector variant, as it performed best in the two first experiments. The performance is evaluated using the Area under the ROC Curve or AUC \cite{hanley1982meaning}, the same metric used in the benchmark baselines of \citet{grnarova2016neural}.

\subsubsection{Baselines}

\citet{grnarova2016neural} devise a model using a word-level CNN for words within a sentence, and a sentence-level CNN processing each sentence sequentially. They append the information about the corresponding medical report category as a vector to each sentence embedding.

They compare their CNN model to two baselines. The first one is the LDA-based Retrospective Topic Model \cite{ghassemi2014unfolding}, a linear kernel SVM trained on the per-report topic distributions. This model is the state-of-the-art model for mortality prediction in the MIMIC-II dataset \cite{saeed2011multiparameter}. The second one is a linear SVM trained on doc2vec \cite{le2014distributed} representations of the reports.

\subsubsection{Target Replication}

The authors also use target replication \cite{lipton2015learning, dai2015semi}. The intuition behind target replication is that the model learns better by replicating the loss at intermediate steps.

More formally, we add to the learning objective a cross entropy term $\ell(y_d^\star, y_d^i| \vec{h_i})$ for every intermediate step $i$ of a training sample $d$, where $y_d^\star$ is the label associated to the corresponding training sample, $y_d^i$ is the predicted label at the intermediate step $i$, and $h_i$ is the hidden state of step $i$ from which a softmax probability is computed. The loss function $\mathcal{L}_d$ for the training sample $d$ becomes:

\begin{equation}
\mathcal{L}_d = \ell(y_d^\star, y_d|\vec{h}) + \frac{\lambda}{n} \sum_{i = 1}^{n} \ell(y_d^\star, y_d^i| \vec{h_i})
\label{eq10}
\end{equation}

In Equation \ref{eq10}, $\lambda$ is a regularization parameter, $y_d$ is the label predicted using the hidden state $\vec{h}$ of the training sample $d$.

In the CNN model, target replication means predicting at each sentence and computing the corresponding loss. However, in our \textit{Structure Tree-LSTM}, it means that we replicate the loss at \textit{hierarchy levels}: we can predict at all units one level below the root (the root's children), two levels (the children of the root's children), or more. Here, we use 1-level target replication. Intuitively, we are replicating the loss at the \textit{report nodes}.

\subsubsection{Mortality Prediction Results}

Our results are reported in Table \ref{exp45}. Our model only came 0.005 short of the CNN baseline, and beat the other two baselines. This can be explained the difference between the CNN baseline and our \textit{Structure Tree-LSTM} model: whereas the CNN baseline processes a patient's reports in the temporal order in which they were issued by the hospital, our model does not incorporate this temporal information. This is important as there is a difference between a good health report coming after a bad health report (recovering patient) and the reverse situation (worsening health). More generally, the children of a Child-Sum Tree-LSTM unit are not processed in any sequential order, and as such sequential order is not preserved. Given that our model nonetheless gives competitive results without temporal information, our future work will focus on modelling sequential order in Child-Sum Tree-LSTMs. It also indicates that our model can efficiently model dependency relations other than structure, such as a group of reports on the same patient. 

Additional examples of applications of this ability include author or user modelling, using an author's writings as the children of the root Tree-LSTM unit. Here, the dependency relation would be authorship.



\section{Conclusions}
\label{5}

To the best of our knowledge, the \textit{Structure Tree-LSTM} is the first attempt to use Tree-LSTMs for texts larger than sentences. We show that our proposed structure-aware document encoders -- the zero-vector variant -- applies attention to all document structural levels. We apply the method for document classification and obtain an average of 9.00\% relative improvement in macro-F1 score with respect to the the best baseline score. 
We also show that our model's hierarchical attention mechanism can be visualised, thereby making the predictions interpretable.

We also test this model on the MIMIC-III dataset, by modelling a patient as the root unit of the Tree-LSTM, and the corresponding medical reports as the children units. We obtain comparable results to the state of the art, coming only 0.005 short in AUC score. We hypothesize that the difference is because the Tree-LSTM cannot encode temporal order, but we note that it successfully modelled structure larger than a single document. This ability could have multiple practical applications, such as modelling people based on their writings.

Finally, we publish a novel document classification dataset of structured Wikipedia articles and release our code to encourage further research on long document encoders.


\bibliography{naaclhlt2019}
\bibliographystyle{acl_natbib}

\appendix

\section{Wikipedia Dataset Details}
\label{sec:supplemental}

The Wikipedia dataset is collected from the English Wikipedia dump of February 1st, 2018\footnote{All Wikipedia dumps are freely available at \url{https://dumps.wikimedia.org/backup-index.html}}. We use a slightly modified version of the WikiExtractor\footnote{The code used is available online at \url{https://github.com/attardi/wikiextractor}} to extract article from the unzipped \texttt{.xml} file. 

The articles are filtered to have a certain length. To do so, we first compute the number of sentences, paragraphs and sections for each article. This is to get an idea of the average length of Wikipedia articles, and then to set a limit on them to filter out stubs. We check percentiles and decide to filter at 25\% to get stubs out of our dataset. Practically, this corresponds to filtering out articles with less than 2 sections, 3 paragraphs and 5 sentences.

Afterwards, we get articles such that they belong to exactly one of 24 categories, and the numbers are detailed in Table \ref{wikitable}. We determine the categories in the table by looking at keywords from the Wikipedia-tagged categories, not the articles themselves, and these categories are excluded from the body of the articles. The resulting dataset has 494,657 articles, and is released as an open resource\footnote{Dataset link to be added in the camera-ready version.}.

\begin{table}[t!]
\begin{center}
\begin{tabular}{|l|r|}
\hline \bf Category & \bf Number of articles \\ \hline
Actors & 28 007 \\
Actresses & 22 208 \\
Aircraft & 12 278 \\
Airlines & 2 496 \\
Artists & 39 618 \\
Cities & 27 090 \\
Comedy & 14 680 \\
Directors & 19 218 \\
Documentaries & 3 848 \\
Drama & 20 523 \\
Footballers & 69 151 \\
Horror & 4 875 \\
Journalists & 15 363 \\
Languages & 6 779 \\
Military Personnel & 17 910 \\
Musicians & 17 603 \\
Novelists & 14 964 \\
Novels & 25 247 \\
Political Parties & 4 233 \\
Politicians & 56 130 \\
Singers & 17 055 \\
Television & 33 434 \\
Video Games & 20 059 \\
Wars & 1 888 \\ \hline
\bf Total & \bf 494 657 \\
\hline
\end{tabular}
\end{center}
\caption{\label{wikitable} Number of articles per category in the Wikipedia dataset used in the first experiment. }
\end{table}

\section{Training Details and Hyperparameters}
\label{hyper}

For all experiments, we used an Adam optimizer with a weight decay of $1\mathrm{e}{-4}$ and a learning rate of $1\mathrm{e}{-2}$. The batch size and hidden layer dimensions are respectively 64 and 128 for the sent2vec-based experiments, and 32 and 64 for the word2vec-based experiments. All models were trained using PyTorch.

We use 300-dimensional \textit{word2vec} model pre-trained on the Google News Corpus\footnote{Available at: \url{https://code.google.com/archive/p/word2vec/}}, and the 700-dimensional \textit{sent2vec}\footnote{Available at: \url{https://github.com/epfml/sent2vec}} model pre-trained on Wikipedia.

\end{document}